\documentclass[10pt,conference,a4paper]{IEEEtran}

\usepackage{amsopn}
\DeclareMathOperator*{\argmax}{arg\,max}

\ifCLASSINFOpdf

\usepackage{graphicx}
\else

\fi

\usepackage{url}

\usepackage{graphics} 
\usepackage{epsfig} 
\usepackage{amsmath} 
\usepackage{amssymb}  

\begin{document}

\title{Fast Gesture Recognition with Multiple Stream Discrete HMMs on 3D Skeletons}

\author{
	\IEEEauthorblockN{Guido Borghi, Roberto Vezzani and Rita Cucchiara}
	\IEEEauthorblockA{DIEF - University of Modena and Reggio Emilia\\ Via P. Vivarelli 10, 41125 Modena, Italy \\
   Email: \{name.surname\}@unimore.it}
}

\maketitle

\begin{abstract}
HMMs are widely used in action and gesture recognition due to their implementation simplicity, low computational requirement, scalability and high parallelism. They have worth performance even with a limited training set. All these characteristics are hard to find together in other even more accurate methods.
In this paper, we propose a novel double-stage classification approach, based on Multiple Stream Discrete Hidden Markov Models (MSD-HMM) and 3D skeleton joint data, able to reach high performances maintaining all advantages listed above. The approach allows both to quickly classify pre-segmented gestures (offline classification), and to perform temporal segmentation on streams of gestures (online classification) faster than real time. We test our system on three public datasets, \textit{MSRAction3D}, \textit{UTKinect-Action} and \textit{MSRDailyAction}, and on a new dataset, \textit{Kinteract Dataset}, explicitly created for Human Computer Interaction (HCI). We obtain state of the art performances on all of them.  

\end{abstract}

\section{Introduction}\label{sec:introduction}

Capturing and understanding people behavior through vision and RGB-D data is a popular and challenging research topic. Since the area is very broad, the proposed systems are often specialized, according with the context, the specific application goals and the type of observed human activity. Depending on the task, the temporal granularity and the semantic abstraction level, different terms have been adopted to describe atomic body movements such as posture, gesture, action, interaction, behavior, and so on.
In this paper, we focus on the recognition of \textit{dynamic body gestures for explicit Human-Computer Interaction} (HCI), that can be defined as follows:

\begin{itemize}
\item \textbf{Dynamic}: the target gesture requires a movement; thus we neglect static postures (e.g., \textit{sitting}, \textit{reading a book});
\item \textbf{Body}: the target gesture is potentially performed using the whole body; thus we discard too local gestures such as finger movements or facial expressions;
\item \textbf{Gestures}: a gesture is a well-defined and time-limited body movement; continuous actions such as \textit{running}, \textit{walking} are not considered;
\item \textbf{Explicit HCI}: we focus on gestures provided by a user which has spontaneously decided to interact with the system; thus, the gesture recognition subsumes a corresponding reaction or feedback at the end of each gesture.
\end{itemize}

Different methods have been proposed in the past to address this problem; currently, the most common solutions adopted in real-time applications include a 3D sensor and a classifier based on Dynamic Time Warping (DTW) or a Hidden Markov Model (HMM) working on body joint positions. 
Even if more complex solutions have been proposed, DTW and HMM based systems have a wider diffusion thanks to their simplicity, low computational requirements and scalability; moreover, even a limited training set allows to reach worth performance. 
Although the recent release of cheap 3D sensors, like \textit{Microsoft Kinect}, bring up new opportunities for detecting stable body points, developing a gesture recognition solution with characteristics of efficiency, efficacy and simplicity together is still far to be accomplished.  Here we propose a new framework, which performs an on-line  \textit{double-stage Multiple Stream Discrete Hidden Markov Model} (double-stage MSD-HMM) to classify gestures.  
In the HCI context we address, users are in front of the acquisition and have the (intentional) need to exchange information with the system through their natural body language. Differently from other works that use standard HMMs and are focused on defining and detecting \textit{ad hoc} feature sets to improve performances, we focus our attention on the whole pattern recognition flow, from the new HMM architectural setting to its implementation details, in order to obtain a real time framework with good performances on speed and classification. 


\section{Related works}
Thanks to the spreading of low cost 3D sensors, the research on  gesture recognition  from RGB-D data has grown interest. In particular, the availability of  almost accurate streams of body joint 3D positions \cite{Shotton11}  allows the body posture or gesture recognition directly from skeleton data, without the need of working on the source depth map. 3D joint position could be affected by noise and estimation errors in presence of occlusions and depth maps intrinsically include richer information. Neural networks and deep architectures are able to extract good features and to correctly classify actions and gestures from the depth maps with impressive performances, when large annotated datasets are available \cite{WangLGZTO15}. Hybrid methods merge skeleton and depth features, in order to get both advantages and increase system performances. To this aim, solutions based on  Random Forest \cite{Zhu_2013_CVPR_Workshops} or multi-kernel learning in \cite{Althloothi2014,Wu12} have been proposed. However, although methods based on skeleton only may be affected by errors and noise, their computational requirements are usually less demanding and more suitable for real-time systems. Thus, in this work we focus on a solution based on 3D joint positions only.
Following the same assumption, Evangedilis \textit{et al.} in \cite{Evangelidis-ICPR-2014} proposes a local, compact and view-invariant skeletal feature to encode \textit{skeletal quad}. Chaudhry \textit{et al.} in \cite{Chaudhry_2013_CVPR_Workshops} proposes a bio-inspired method to create 3D discriminative skeletal feature. In \cite{Eweiwi2015} action recognition is performed through skeletal pose-based features, built on location, velocity and correlation joint data. Xia \textit{et al.} in \cite{xia2012view} uses histograms of 3D joints locations to perform classification through discrete HMMs. HMM models has been widely used due to their low computational load and parallelization capabilities. Also \cite{Wu2014} suggests an approach based on HMMs by modelling the state emission probability with neural networks, which somehow nullify the advantages of HMM models. 
Finally, very few works \cite{AleksicK06,08sch16} have been proposed in the past with discrete weighted multiple stream HMMs, but only in facial expression and handwriting recognition tasks respectively (no gesture recognition task). 

\section{Offline Gesture Classification}
\label{sec:offline}
The core part of the proposal is a probabilistic solution for the classification problem of a pre-segmented clip, containing a single gesture (offline classification).
Given a set of $C$ gesture classes $\Lambda={\lambda^1 \ldots \lambda^C}$, we aim at finding the class $\lambda^*$ which maximizes the probability $P(\lambda|O)$, where $O=\{o_1 \ldots o_T\}$ is the entire sequence of frame-wise observations (i.e. the features). 
A set of HMMs, each trained on a specific gesture, is the typical solution adopted for this kind of classification problems \cite{rabiner}. The classification of an observation sequence $O$ is carried out selecting the model $\lambda^*$  whose likelihood is highest. If the classes are a-priori equally likely, this solution is optimal also in a Bayesian sense.  

\begin{equation}
\lambda^* = \argmax_{1 \le c \le C} \left[ P \left( O|\lambda^c\ \right) \right] 
\label{eq:argmax}
\end{equation}

If the decoding of the internal state sequence is not required, the standard recursive forward algorithm for HMMs with the three well known initialization, induction and termination equations can be applied:
\begin{equation}
\begin{array}{rcl}
\alpha_1(j) & = & \pi_i b_j(o_1), 1 \le i \le N    \\
\alpha_{t+1}(j) & =&  \left[ \sum_{i=1}^N {\alpha_t(i) a_{ij}} \right] b_j\left( o_{t+1} \right) \\
P \left(O| \lambda \right) & =&  \sum_{j=1}^N {\alpha_T \left(j \right)}  
\end{array}
\label{eq:HMMequations}
\end{equation}
where $\pi = p(q_i=s_i)$ is the initial state distribution, $a \in A$, matrix A describes the transition $p(q_t=s_i|q_{t-1} = s_i)$ for hidden states $S$ ($q_t$ is the current state), $b_j(o)$ depends on the type of the observation and  defines the emission state probabilities.\\
A common solution is based on Gaussian Mixture Models (GMMs) learned on a feature set composed of a set of continuous values \cite{VezzaniIcip09}. The term $b_j(o_t)$ of Eq. \ref{eq:HMMequations} would be approximated as:
\begin{equation}
b_j(o_t)=\sum\limits_{l=1}^M{c_{jl}\mathcal{N}\left( o_t \vert \mu_{jl},\Sigma_{jl} \right) }
\label{eq:bjCont}
\end{equation}
\noindent where $M$ is the number of Gaussian components per state, $\mu_{jl}$ and $\Sigma_{jl}$ are the Gaussian parameters, and $c_jl$ are the mixture weights.

\subsection{Feature set}\label{sec:feature set}
We assume to already have the body joint 3D coordinates as input \cite{Shotton11}. In this work, we exploited a simple feature set directly derived from the joint stream, discriminative enough to obtain reasonable classification rates. Additional features may be included without changing the classification schema, but this may go to the detriment of the computational complexity and to the overall efficacy due to the curse of dimensionality. 
Thus, only nine features are extracted for each selected body joint $K_i$. Given the sequence of 3D positions of the joint $K_i^t=\left(x_i^t, y_i^t, z_i^t\right)$, we define $o_t(i) = \left\{ o^1_t(i) \ldots o^{9}_t (i) \right\}$ as:

\begingroup\makeatletter\def\f@size{9}\check@mathfonts
\begin{equation}
\begin{array}{lcc}
o^1_t(i), o^2_t(i), o^3_t(i) = x_i^t - \hat{x}_i^t, y_i^t - \hat{y}_i^t, z_i^t - \hat{z}_i^t \\
o^4_t(i), o^5_t(i), o^6_t(i) = x_i^t - x_i^{(t-1)}, y_i^t - y_i^{(t-1)}, z_i^t - z_i^{(t-1)}  \\

o^7_t(i), o^8_t(i), o^9_t(i) = x_i^t - 2x_i^{(t-1)} +x_i^{(t-2)},\\
\ \ \ \ y_i^t-2y_i^{(t-1)} +y_i^{(t-2)}, z_i^t - 2z_i^{(t-1)} +z_i^{(t-2)}  \\
\end{array}
\label{eq:featureset}
\end{equation}
\endgroup

\noindent where $\{o^1 \ldots o^3\}$ are the joint positions with respect to a reference joint $\hat{k}$ selected to make the feature set position invariant; $\{o^4 \ldots o^6\}$ and $\{o^7 \ldots o^9\}$ are the speed and acceleration components of the joint respectively. This approach is inspired by \cite{Eweiwi2015,Zhu_2013_CVPR_Workshops}. A linear normalization is applied to normalize the feature values to the range of $[-1, 1]$. Selecting a subset (or the whole set) of $G$ body joints, the complete feature set $o_t$ for the $t-th$ frame is obtained as a concatenation of $D=9 \cdot G$ features. Thanks to the limited dependencies among the features, fast and parallel computation is allowed.

\subsection{Multiple Stream Discrete HMM}
The computation of exponential terms included in Eq. \ref{eq:bjCont} is time-consuming. Moreover, the Gaussian parameters $\mu_{jl},\Sigma_{jl}$ may lead to degenerate cases when learned from few examples. In particular, in the case of constant features the corresponding $\Sigma_{jl}$ values becomes zero. Even if some practical tricks have been proposed in these cases \cite{rabiner}, overall performances generally decrease when few input examples are provided in the learning stage. For these reasons, we propose to adopt discrete HMM. All continuous observations $O$ are linearly quantized to $L$ discrete values $l \in [1 \ldots L]$.
The adoption of discrete distributions to model the output probability solves the above mentioned numerical issues. In addition, to improve the generalization capabilities of HMM with limited samples for each state, we adopted a set of $D$ independent distributions --one for each feature item-- called \textit{streams}; the term $b_j(o_t)$ of Eq. \ref{eq:bjCont} is replaced with the following one:
\begin{equation}
b_j(o_t)=\prod\limits_{d=1}^D { \left( h_j^d(o_t^d)  \right) ^ {\alpha_d} }
\label{eq:bjMS}
\end{equation}

\noindent where $\left( h_j^d(o_t^d)  \right) $ is the emission probability of the discrete value $o_t^d$ in the $j-th$ state and $\alpha_d$ are weighting terms. 
The observation of each stream is thus \textit{statistically independent} from the others given the current state $q_t$.\\
The weight coefficients $\alpha_d$ may be used to take into account the different  classification capability of each joint and each feature component (see Eq. \ref{eq:featureset}).

\begin{figure*}[th!]
    \centering
    \includegraphics[width=0.85\textwidth]{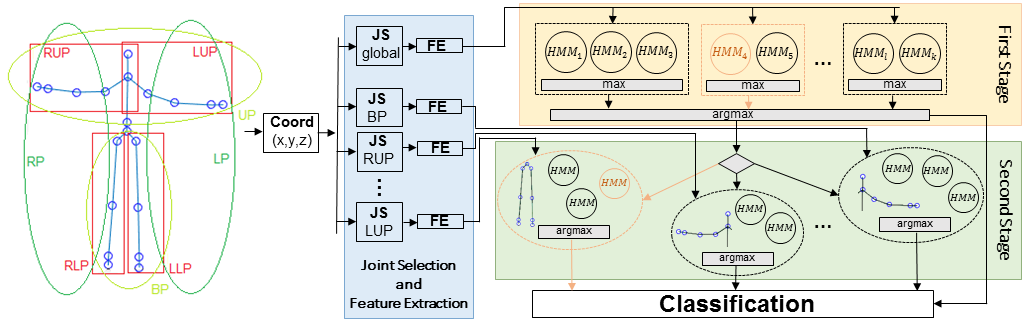}
    \caption{Block-diagram overview of the proposed method.}
    \label{fig:general}
\end{figure*}

\section{Double-stage classification}
\label{sec:doubleStage}
The previous described approach does not discriminate if there are some parts of the body that are more significant for a given gesture, since the same feature set is provided as input to all the HMMs. For example, the gesture ``hello'' can be better recognized if the hand's joints are analyzed alone without other distracting joints. However, the responses HMM with different feature sets becomes not comparable. 
Following this observation,  a \textit{double-stage} classification system based on MSD-HMM is proposed as outlined in Fig. \ref{fig:general}. Gestures are grouped into sets depending on the sub set of the interested body joints.
The first classification stage recognizes the gesture group (i.e., discovers which part of human body is most involved in the gesture), while the second stage provides the final classification among the gestures in the selected group. The MSD-HMMs of the first stage work on a global subset of joints extracted from the whole body, while the MSD-HMMs of the second stage are more specific on the body part involved. In particular, the global subset of joints (used to compute the features of Eq. \ref{eq:featureset}) contains the left and right foot, the left and right hand and the head. Other joints, like shoulders or knees, are instead used in the second stage.
Gesture groups of the second-stage correspond with the partition reported on the left part of Fig.\ref{fig:general}. Four local-areas (i.e., Right Upper Part (RUP), Left Upper Part (LUP), Right Lower Part (RLP), Left Lower Part (LLP)) are firstly defined. Then, four additional macro-areas are created as combinations of local-areas (i.e., Upper Part (UP), Bottom Part (BP), Right Part (RP) and Left Part (LP)). Eight gesture clusters are correspondingly defined based on the main body part involved. A different set of stream weights $\alpha_d$ (see Eq. \ref{eq:bjMS}) is computed for each HMM, using the average motion of each joint: the high the motion of a body part in a gesture is, the high is the corresponding stream weights. 

\section{Online Temporal Segmentation and Classification of Gestures}
\label{sec:online}
Differently from offline gesture classification, online recognition requires to estimate the most likely gesture currently performed by the monitored subject, given only the observations until now. The observed sequence may contain more instances and the current gesture may be in progress. The original proposed system is able to detect gestures on-line, overtaking classical static approaches that are characterized by strong prior hypothesis. For example, common \textit{sliding window} methods implicitly apply a strong constraint on the average and maximum duration of each gesture. Since HCI is often characterized by a high variability on the gesture duration and noise, the proposed solution is not based on fixed-length rigid windows. 
Using HMM, the temporal evolution of a gesture is correlated with the hidden state probability. In particular, exploiting the \textit{left-right} transition model \cite{rabiner} the first and last state of the chain can be exploited to detect the temporal boundaries of each gesture. 
The following rules are applied to the first-stage classification layer of Fig. \ref{fig:general}:
\begin{itemize} 
\item \textbf{Beginning detection}: the beginning of a gesture candidate is detected by analyzing the first hidden state of each HMM. The adoption of a left-right transition model is mandatory. The most likely state of HMM will be the first one during non-gesture times, while the following states will be activated once the gesture starts.
A voting mechanism is exploited, counting how many HMMs satisfy the following condition: 
\begingroup\makeatletter\def\f@size{8}\check@mathfonts
\begin{equation}
\phi(HMM_k)= \bigg \{
\begin{array}{rl}
\label{eq:2poll}
0 & \alpha_t(1) = \biggl[\sum_{i=1}^N\alpha_t(i)a_{i1}\biggr] b_1(o_t) \geq th \\
1 & \alpha_t(1) = \biggl[\sum_{i=1}^N\alpha_t(i)a_{i1}\biggr] b_1(o_t) < th \\
\end{array}
\end{equation}
\endgroup
where $N$ is the total number of hidden states in HMMs and $b$ is defined in Eq. \ref{eq:bjMS}. If a large number of models satisfies Eq. \ref{eq:2poll}, a gesture is performing. The threshold $th$ on $\alpha_t(1)$ checks the probability to be still at the instant $t$ in the first hidden state.


\item \textbf{End detection}: if a gesture is currently performed (as detected by the previous rule) the most likely gesture is computed at each frame as in Eq. \ref{eq:argmax}. A probability distribution analysis of the last state of the corresponding HMM is performed in order to detect the end of the gesture:
\begingroup\makeatletter\def\f@size{10}\check@mathfonts
\begin{equation}
\alpha_t(N) = \biggl[\sum_{i=1}^N\alpha_t(i)a_{iN}\biggr] b_j(o_t) \geq th
\end{equation}
\endgroup
\noindent where $N$ is the total number of hidden states and $th$ is a threshold on $\alpha_t(N)$, the probability to be at the instant $t$ in the last ($N$-th) state after observations $O(t_s...t_e)$.


\item \textbf{Reliability Check}: this rule is applied to filter out false candidates (non valid or incomplete gestures, for example). The sequence of hidden state probabilities obtained with the selected HMM model on the observation window is analyzed. Starting from the set $S$ of hidden states, the subset $\hat{S} \subset S$ is defined as follows:

\begin{equation}
\hat{S}= \{ s_j | \forall t \in [ t_s, t_e], \alpha_t(j) \geq th \} 
\end{equation}

\noindent The following checks are performed to validate the candidate gesture:
\begin{equation}
\begin{array}{rcl}
\#\hat{S} & \geq  & \frac{2}{3}N  \\
S_{N-1} & \in  & \hat{S}  \\
\end{array}
\end{equation}

\noindent where $\#\hat{S}$ is the cardinality of the set $\hat{S}$. The previous checks guarantee that at least $\frac{2}{3}$ of hidden states (included the second to last) are visited. A state is defined as visited if there is an instant $t$ characterized by a high probability ($th>0.90$) to be in that state. 
\end{itemize}

\noindent Once the first stage classification layer provides a valid temporal segmentation and the corresponding simultaneous gesture classification, the second-stage layer is exploited to refine the classification.

\begin{figure*}[th!]
    \centering
    \includegraphics[width=0.60\textwidth]{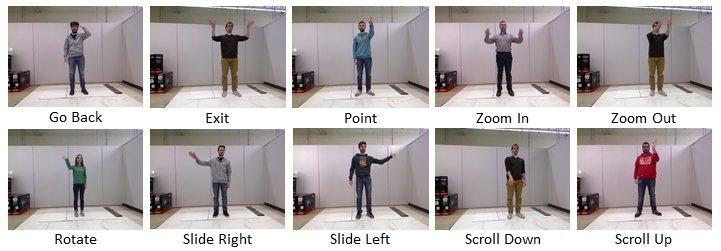}
    \caption{Action classes from our \textit{Kinteract Dataset}. Classes are explicitly created for Human Computer Interaction.}
    \label{fig:kinteract dataset}
\end{figure*}

\section{Experimental Results} \label{sec:results}
We test our system on three public and famous datasets, \textit{MSRAction3D}, \textit{MSRDailyActivity3D} and \textit{UTKinect-Action}, as well as on a new custom dataset (\textit{Kinteract Dataset}) we developed for HCI. 

\subsection{MSRAction3D Dataset}\label{sec:msr}
MSR Action3D dataset \cite{Li10} contains 20 action classes performed by 3 subjects: \textit{high arm wave, horizontal arm wave, hammer, hand catch, forward punch, high throw, draw x, draw tick, draw circle, hand clap, two hand wave, side-boxing, bend, forward kick, side kick, jogging, tennis swing, tennis serve, golf swing and pickup} \& \textit{throw}. In total there are 567 action instances.
This dataset fits our method due to the absence of human-object interaction and it contains only 3D joint position. This is one of the most used dataset for human action recognition task, but in \cite{Padilla-Lopez14} are reported some issues about validation methods: the total number of samples used for training and testing is not clear, because some 3D joint coordinates are null or with highly noise. To avoid ambiguity, the complete list of valid videos we used is publicly available\footnote{\url{http://imagelab.ing.unimore.it/hci}}.  The validation phase has been carried out following the original proposal by Li \textit{et al.} \cite{Li10}. Three tests are performed: in the first two tests, 1/3 and 2/3 of the samples are used for training and the rest for the testing phase; in the third test half of the subjects are used for training (subjects number 1, 3, 5, 7, 9) and the remainder for testing (2, 4, 6, 8, 10).

\subsection{UTKinect-Action Dataset}
UTKinect-Dataset \cite{xia2012view} contains 10 action classes performed by 10 subjects: \textit{walk, sit down, stand up, pick up, carry, throw, push, pull, wave hands, clap hands}; each subject performed every action twice. There are 200 action sequences. This dataset is challenging due to high intra-class variations and the variations in the view point. Moreover, because of the low frame rate and the duration of same actions, some sequences are very short. The validation method proposed in \cite{xia2012view} has been adopted, using a leave one sequence out cross validation. 

\subsection{MSRDailyActivity3D Dataset}\label{sec:daily}
DailyActivity3D dataset \cite{Wu12} contains daily actions captured by a Kinect device. There are 16 activity classes performed twice by 20 subjects: \textit{drink, eat, read book, call cellphone, write on a paper, use laptop, use vacuum cleaner, cheer up, sit still, toss paper, play game, lay down on sofa, walk, play guitar, stand up, sit down}. Generally, each subject performs the same activity standing and sitting on the sofa. There is a total of 320 long activity sequences. 
Since our proposed method is specifically conceived for dynamic gestures (see Sec. \ref{sec:introduction}), still actions (e.g., read book, sit still, play game and lay down on the sofa) have been removed during the evaluation. We follow the cross-subject test proposed in \cite{Wu12,Zanfir}. 

\subsection{Kinteract Dataset}
In addition, we collect a new dataset publicly available$^1$, which has been explicitly designed and created for HCI. Ten common types of HCI gestures have been defined: \textit{zoom in, zoom out, scroll up, scroll down, slide left, slide right, rotate, back, ok, exit}. For example, these gestures can be used to control an application using the \textit{Kinect} sensor as a natural interface. 
Gestures are performed by 10 subjects for a total of 168 instances and are acquired by standing in front of a stationary \textit{Kinect} device. Only the upper part of the body  (from shoulders to hands) is involved. Each gesture is performed with the same arm by all the subjects (despite they are left or right handed). This dataset allows to highlight the advantages of our solution in a real world context. 

\subsection{Experiments}
\begin{figure}[bth]
    \centering
    \includegraphics[width=0.60\columnwidth]{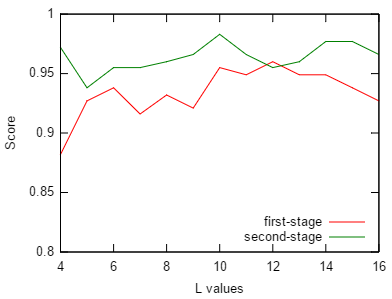}
    \caption{System performance w.r.t. quantization levels}
    \label{fig:quant}
\end{figure}

The output of the first-stage HMM can be directly used for classification (using \textit{argmax} instead \textit{max} in Fig. \ref{fig:general}). We report the corresponding performance as reference for the final double-stage one. 
The number of HMM hidden states is set to 8. The number $L$ of quantization levels (see Sec. \ref{sec:feature set}) has been empirically selected on the MSRAction3D dataset. As reported in Fig. \ref{fig:quant}, the best value is $L=10$. Finally, stream weights are equally initialized to 1 by default.

Table \ref{tab:MsrActionInternal} contains an internal comparison of the system on the  \textit{MSRAction3D} dataset, based on cross subjects evaluation method. The complete system is compared with baseline solutions where the feature normalization (FN) and the stream weight (WMS) steps are not performed.  

\begin{table}[h]
\caption{System performance}
\centering
\begin{tabular}{|c|c|c|}
\hline
\multicolumn{3}{|c|}{\textbf{MSRAction3D}}  \\ \hline \hline
\textbf{System Part} & \textbf{First Stage} & \textbf{Second Stage}    \\ \hline
Base Method              & 0.739       & 0.802          \\ \hline
Base + FN                   & 0.845       & 0.873          \\ \hline
Base + FN + WMS             & 0.861       & \textbf{0.905} \\ \hline
\multicolumn{3}{c}{\footnotesize FN: Feature Normalization; WMS: Weighted Multiple Streams} 
\end{tabular}
\label{tab:MsrActionInternal}

\end{table}
 
Table \ref{tab:msr} reports the performance of state of the art methods on the same dataset. 
Results show that our method is in line with the state of the art, where methods based on skeleton only are listed. The approach \cite{Eweiwi2015} provides better results but is really much more expensive in terms of feature extraction and classification time, as described below.

\begin{table}[]
\caption{Results on the MSRAction3D dataset}
\label{tab:msr}
\centering
\begin{tabular}{|c|c|c|c|}
\hline
\textbf{Methods} &\multicolumn{3}{|c|}{\textbf{Accuracy}}  \\ \hline \hline
					 & \textbf{1/3} & \textbf{2/3}  & \textbf{cross}    \\ \hline
HOJ3D \cite{xia2012view}                 & 0.962       & 0.971          & 0.789    \\ \hline
HMM + DBM \cite{Wu2014}                 & -       & -          & 0.820    \\ \hline
EigenJoints \cite{YangT12}                 & 0.958       & 0.977         & 0.823    \\ \hline
HMM + GMM (our implementation)          & 0.861       & 0.929          & 0.825  \\ \hline
Actionlet Ensemble (skeleton data)    \cite{Wu12}        & -       & -          & 0.882  \\ \hline
Skeletal Quads    \cite{Evangelidis-ICPR-2014}        & -       & -          & 0.898  \\ \hline
LDS \cite{Chaudhry_2013_CVPR_Workshops}                 & -       & -          & 0.900    \\ \hline
Cov3DJ \cite{Hussein2013}                  & -       & -			& 0.905 \\ \hline
Our                  & 0.943       & 0.983			& 0.905 \\ \hline
FSFJ3D \cite{Zhu_2013_CVPR_Workshops}                  & -       & -			& 0.909 \\ \hline
KPLS \cite{Eweiwi2015}                  & -       & -			& 0.923 \\ \hline
\end{tabular}
\end{table}

\begin{figure}[h]
    \centering
    \includegraphics[width=0.60\columnwidth]{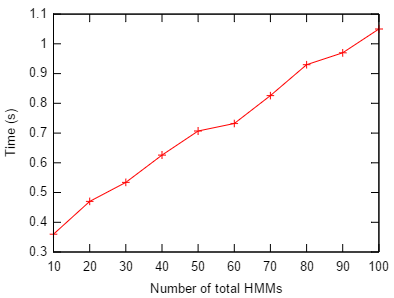}
    \caption{Classification time w.r.t. the number of classes}
    \label{fig:time}
\end{figure}

Results on \textit{UTKinect-Action} dataset are reported in Table \ref{tab:utk}: we run the experiment 20 times and we get the mean performance; however, the best accuracy obtained in a run is 0.965. Table \ref{tab:msrdaily} reports results for the\textit{ MSRDailyAction} dataset. Even if the comparison is not completely fair since we considered a subset of gestures and we only use skeleton data, the reported performance confirms the generalization capability of our method as well as its efficacy on long and generic actions. The proposed system has an overall accuracy of 0.974 on the \textit{Kinteract} dataset performing a cross subject test.
Finally, we test the online temporal segmentation. Instances belonging to \textit{MSRAction3D} and \textit{Kinteract} dataset are merged in a continuous stream. Corresponding results are reported in Table \ref{tab:online}.
\begin{table}
\caption{Results on the UTKinect-Action dataset}
\label{tab:utk}
\centering
\begin{tabular}{|c|c|}
\hline
\textbf{Methods} &\multicolumn{1}{|c|}{\textbf{Accuracy}}  \\ \hline \hline
DSTIP+DCSF\cite{Zhu_2013_CVPR_Workshops}                    & 0.858 \\ \hline
FSFJ3D (skeleton data)\cite{Zhu_2013_CVPR_Workshops}                    & 0.879          \\ \hline
SNV \cite{Yang2014}                  & 0.889      \\ \hline
Our                  & 0.895       \\ \hline
HOJ3D \cite{xia2012view} & 0.909 \\ \hline
\end{tabular}
\end{table}
The threshold for state analysis probability distribution is $th=0.9$ (see eq. \ref{eq:2poll}). A temporal segment is valid only if has a Intersection over Union (IoU) with ground truth instance larger than overlap threshold ($\sigma=0.5$); the ground truth instance with greater IoU defines the current class action that has to be classified. Results show that our original temporal segmentation method can be used in a real world framework.
We implemented the proposed system in C++ and tested on a computer with Intel i7-4790 (3.60GHz) and 16 GB of RAM. The framework is able to extract and calculate features, perform feature quantization, evaluate stream weights and classify a single action with an average time of $4.4 \cdot 10^{-2}s$ (tested on the \textit{MSRAction3D} dataset with 20 HMM for each stage). Performance results with respect to the number of HMMs (for each stage) are reported in Fig. \ref{fig:time}. The total number of HMM of each stage corresponds to the potential number of recognizable gesture classes. Results are normalized by the mean gesture length (41 frames). The complete online system runs at about 80.35 frames per second when trained on the \textit{MSRAction3D} dataset. Its exploitation on real time systems is then guarantee, differently to other recent state of the art solutions \cite{BMVC.23.124:abbreviated,Fanello2013}; in particular, the feature extraction and classification time for a single action in \cite{Eweiwi2015} is about 300 ms.

\begin{table}[!h]
\caption{Results on the MSRDailyAction dataset}
\label{tab:msrdaily}
\centering
\begin{tabular}{|c|c|}
\hline
\textbf{Methods} &\multicolumn{1}{|c|}{\textbf{Accuracy}}  \\ \hline \hline
DTW  \cite{Muller2006}        & 0.540          \\ \hline
Moving Pose   \cite{Zanfir}              & 0.738          \\ \hline
HON4D \cite{Oreifej13hon4d}                  & 0.800       \\ \hline
Our                  & 0.833       \\ \hline
Actionlet Ensemble    \cite{Wu12}              & 0.845          \\ \hline
\end{tabular}
\end{table}


\begin{table}[]
\caption{Results for Online Temporal Segmentation}
\label{tab:online}
\centering
\begin{tabular}{|c|c|c|}
\hline
\textbf{Dataset} &\ \textbf{Detection Rate} &   \textbf{Recognition Rate}\\ \hline \hline
MSRAction3D                  & 0.782   & 0.871  \\ \hline
Kinteract                  & 0.892    & 0.943  \\ \hline
\end{tabular}
\end{table}
 
 \section{Conclusions}
In this paper, we investigate and improve the use of on-line double-stage Multiple Stream Discrete HMM (MSD-HMM), that are widely used due to their implementation simplicity, low computational requirements, scalability and high parallelism, in action and gesture recognition fields. We also demonstrate that HMMs can be successfully used for gesture classification tasks with worth performances even with a limited training set.
Thanks to a double-stage classification based on MSD-HMM, our system allows both to quickly classify and perform online temporal segmentation, with a great generalization capability. Results are in line with the state of the art on three public and challenging datasets.
Before our method HMM were not able to compete in gesture classification task with other state-of-the-art methods. 
Moreover, a new gesture dataset is introduced, namely \textit{Kinteract dataset}, which is explicitly designed and created for HCI. 
 
\bibliographystyle{IEEEtran}
\bibliography{biblio}

\end{document}